\documentclass[sigconf]{acmart}
\usepackage{multirow}
\AtBeginDocument{%
  }

\setcopyright{acmlicensed}
\copyrightyear{2024}
\acmYear{2024}
\acmDOI{XXXXXXX.XXXXXXX}

\begin{document}

\title{Demo2Vec: Learning Region Embedding with Demographic Information}

\author{Ya Wen}
\affiliation{%
  \institution{The University of Hong Kong}
  \city{Hong Kong}
  \country{China}
}
\email{wenya@connect.hku.hk}

\author{Yulun Zhou}
\authornote{Corresponding author.}
\affiliation{%
  \institution{The University of Hong Kong}
  \city{Hong Kong}
  \country{China}
}
\email{yulunzhou@hku.hk}

\renewcommand{\shortauthors}{Wen and Zhou}

\begin{abstract}
  Demographic data, such as income, education level, and employment rate, contain valuable information of urban regions, yet few studies have integrated demographic information to generate region embedding. In this study, we show how the simple and easy-to-access demographic data can improve the quality of state-of-the-art region embedding and provide better predictive performances in urban areas across three common urban tasks, namely check-in prediction, crime rate prediction, and house price prediction. We find that existing pre-train methods based on KL divergence are potentially biased towards mobility information and propose to use Jenson-Shannon divergence as a more appropriate loss function for multi-view representation learning. Experimental results from both New York and Chicago show that \textit{mobility + income} is the best pre-train data combination, providing up to 10.22\% better predictive performances than existing models. Considering that mobility big data can be hardly accessible in many developing cities, we suggest  \textit{geographic proximity + income} to be a simple but effective data combination for region embedding pre-training.      
\end{abstract}




\maketitle

\section{Introduction}

Learning region embedding is one of the most fundamental challenges in enabling transferable urban prediction models. Region embedding serves as a condensed representation of the geographical and social context of locations. When effectively learned, region embedding has the potential to forecast urban trends across various tasks and even in different cities. 

The choice of input data for creating region embedding significantly impacts the quality of the embedding. Quality, in this context, refers to how well the region embedding performs in predicting urban outcomes in various cities. Previous research has utilized different input data to generate region embedding, resulting in diverse urban prediction performances. For example, proximity measures have been commonly used to capture spatial similarities between locations \cite{yanluo,HRE}. Urban mobility data is another frequently employed input that has shown high accuracy in urban prediction tasks like check-in prediction \cite{zechenli} and land use classification \cite{shuangbinwu}. Researchers such as Li et al. and Zhang et al. have leveraged Point-of-Interest (POI) data for region embedding \cite{zechenli,mingyangzhang}. Despite the inclusion of various types of information in region embedding, certain essential urban characteristics, particularly demographic data, have been overlooked.

Demographic information is among the most fundamental characteristics of urban regions and is very easily accessible thanks to regular government census. Extensive urban studies have reported strong associations between various demographic attributes and urban dynamics. For example, the crime rate is reported to be strongly associated with regional income, especially in western countries \cite{crimeincome}. Additionally, urban segregation \cite{incomesegregation}, where individuals of varying income levels utilize urban spaces differently, has established a theoretical basis for understanding the connection between demographic traits and urban dynamics. However, most existing studies on region embedding focus on "big" datasets, paying insufficient attention to "small", classic, and potentially meaningful datasets such as demographic information. 

In this study, we examine the possibility and effectiveness of integrating demographic information in learning region embedding. We first evaluate the performance of incorporating income, one representative demographic feature, in improving the predictive performance on three downstream tasks across New York and Chicago compared with state-of-the-art model performances. Then we extend the examination to other demographic information such as age, education level, and employment rate. 

Our contributions can be summarized as follows: 1. We propose the use of Jenson-Shannon (JS) Divergence as a more effective loss function for multi-view representation learning of urban region embedding. 2. We report that regional income information is effective in improving regional embedding learning performance by up to 10.22\%. 3. The effectiveness of income information and other demographic attributes is validated across three tasks in two cities.

\section{Methods}

\subsection{\textbf{Multi-view Representation Learning}}

An urban area is divided into \(n\) non-overlapping regions and the representation learning involves generating a low-dimensional embedding for each region, i.e., \(\mathcal{E} = \{\vec{e}_1, \vec{e}_2, \ldots, \vec{e}_n\}, \vec{e} \in \mathbb{R}^d, \forall \vec{e} \in \mathcal{E}\), where d is the embedding size. The learned embedding can then be applied to various urban prediction and classification tasks. Multi-view graph-based learning efficiently integrates region correlations from multiple data sources and achieves satisfactory performance \cite{Fu,mingyangzhang}. Graph-based methods construct region graphs where nodes represent distinct regions, and a set of edge types captures the correlation between regions in different aspects. Source and target edges derived from human mobility data, POI edges, and geographic neighbor edges are the common edge types for graph formation. 

We use Heterogeneous Region Embedding (HRE) \cite{HRE}, a state-of-the-art module for effective fusion of multi-source data and representation learning. The module consists of a relation-aware GCN that introduces edge embedding, a self-attention layer for sharing between edge-specific region embedding, and an attention-based fusion layer to finalize the region embedding. A multi-task learning framework is utilised to train the model with loss functions defined specifically for each pre-training dimension. Intuitively, regions are more likely to be similar to nearby regions. Therefore, we form the geographic neighbor loss \( L_{n} \) as a triplet loss guiding the model to map adjacent regions closer and push non-neighboring regions farther away in the embedding space. POI loss \( L_{poi} \) is defined to minimize the error between the real POI similarity matrix and the one reconstructed from the learned embeddings. KL divergence is adopted to calculate the mobility loss  \( L_{mobility} \) by minimizing the difference between the real-world trip  distribution and the predicted distributions from the corresponding source and target region embeddings. As a result, the final objective function is formulated as: 
\begin{displaymath}
  \mathcal{L} = L_{\text{n}} + L_{\text{poi}} + L_{\text{mobility}}
\end{displaymath}

When demographic information is integrated to learn region embedding, the loss function is expanded as, 

\begin{displaymath}
  \mathcal{L} = L_{\text{n}} + L_{\text{poi}} + L_{\text{mobility}} + L_{\text{demo}}
\end{displaymath}

,where \( L_{demo} \) is a loss function term for demographic similarity encoded below. 

\subsection{\textbf{Encoding Demographic Information}}
\begin{figure}
    \centering
    \includegraphics[width=1\linewidth]{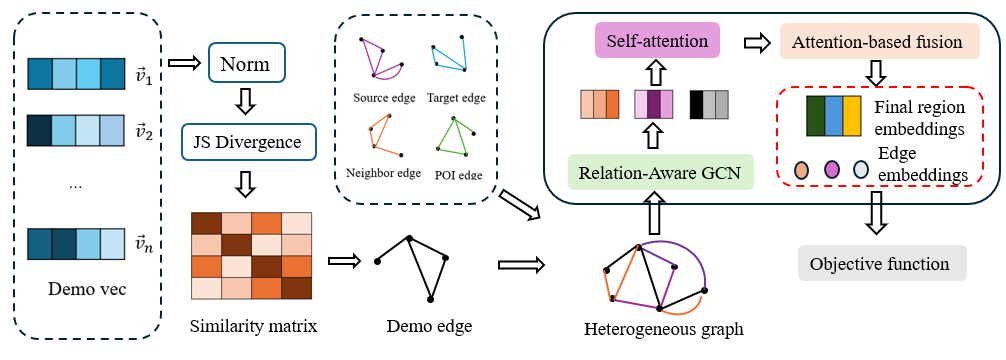}
    \caption{Demographic information encoding and the model structure.}
    \label{fig:model}
\end{figure}

Most demographic information contains a distribution of values, where each regional sum is allocated into multiple categories, with figures indicating the number of citizens belonging to each category. For example, the American Community Survey split the household income into 10 levels in NYC. As shown in Figure \ref{fig:model}, we encode the population distributions as vectors and perform normalization. Then, we use JS divergence to quantify the similarity between the two region distributions and generate corresponding similarity matrices. Afterwards, each region is connected with its top \(k\) similar regions in the heterogeneous graph. The HRE module then utilises the connectivity relationships in a multi-edge setting and generates unified region embeddings for downstream tasks and edge embeddings for loss function calculations.

\subparagraph{\textbf{Jenson-Shannon Divergence.} } 
In multi-task learning, the balance in the loss contribution from each task is critical for effective learning. In existing literature, the mobility loss function is measured by KL divergence. Since KL divergence is unbounded and asymmetric, i.e., \(D_{\text{KL}}(P \parallel Q) \neq D_{\text{KL}}(Q \parallel P)\). As a result, we find that the mobility loss can be several orders of magnitude larger than other pre-training dimensions, dominating the learning process and leaving other dimensions insufficiently learned. Therefore, we choose JS Divergence\cite{JSD} to compute the loss for dimensions with a value distribution, including mobility and demographics. With \(p,q\) being two distributions, JS Divergence is defined by:
\[
D_{\text{JS}}(p \parallel q) = \frac{1}{2} D_{\text{KL}}\left(p \parallel \frac{p+q}{2}\right) + \frac{1}{2} D_{\text{KL}}\left(q \parallel \frac{p+q}{2}\right)
\]
where \(D_{\text{KL}}\) is the Kullback–Leibler (KL) divergence, given by:
\[
 D_{\text{KL}}(P \parallel Q) = \sum_{i} P(i) \log \frac{P(i)}{Q(i)}
\]
JS Divergence is symmetric and ranges from 0 to 1. Consequently, it generates comparable loss values for all pertaining dimensions, leading to a more stable training process and full utilisation of multi-view input data. 

\section{Experiments}
\subparagraph{\textbf{Datasets and Cities}.}
We collect real-world urban datasets of region division, demographic data, POI data, for-hail vehicles, such as Uber and Lyft trip records, check-in, crime and house price data for two major cities in the United States: New York City (NYC) and Chicago (CHI). The detailed statistics and sources of the dataset are shown in Table \ref{tab:datasource}. We study the urban region dynamics at the Neighborhood Tabulation Area (NTA) and Community Area (CA) levels for NYC and CHI, respectively.  We gather demographic data from the US Census Bureau\cite{uscensus}, including a broad range of essential socioeconomic indicators such as household income, age, education level, occupation, and foreign-born population.

\begin{table}[]
\caption{Dataset sources and descriptive statistics.}
\label{tab:datasource}
\begin{tabular}{llll}
\hline
\textbf{Dataset}       & \multicolumn{1}{c}{\textbf{NYC}} & \multicolumn{1}{c}{\textbf{CHI}} & \multicolumn{1}{c}{\textbf{Source}} \\ \hline
Regions                & 192                              & 77                               & Census Bureau \cite{uscensus} \\ \hline
Demographics       & -                                & -                                &            Census Bureau \cite{uscensus}                        \\
 Trips &                              21,046,467&                                  8,021,882&  TLC\cite{nyctrip}, TNP\cite{chicago_transportation}\\
POI                &                         62,451&                                  31,574&     NYCOD\cite{nyc_opendata}, CHIDP\cite{chicago_dataportal}\\ \hline
Check-in           &                                  918,873&                   320,920&              Foursquare \cite{Foursquare}                       \\
Crime              &                                  478,571&                                  273,097&              NYCOD\cite{nyc_opendata}, CHIDP\cite{chicago_dataportal}\\
House price        &                                  54,104&                                  44,448& Zillow\cite{Zillow}, NYCDOF\cite{nyc_finance}\\ \hline
\end{tabular}
\end{table}

\subparagraph{\textbf{Downstream Tasks.}}
We evaluate the performance of \textbf{Demo2Vec} across three downstream tasks. 
\begin{itemize}
    \item Check-in amount, which reflects the region's popularity.
    \item Crime rate, which is the crime count per 10,000 population.
    \item Median house price with a unit of dollar per square foot.
\end{itemize}
For each task, we measure the predictive performance using region embedding as the sole input. We apply a simple Ridge regression model to conduct the prediction and perform k-Fold cross-validation, where \(k=5\). Model evaluation metrics include mean absolute error (MAE), root mean squared error (RMSE), and coefficient of determination ($R^2$). 


\subparagraph{\textbf{Experimental Design.}}
We explore various combinations of pre-training dimensions to assess their individual contributions.  For every combination, we first tune parameters such as the learning rate and weight decay to find the optimal settings. Then, we conduct 10 runs of the model for each combination and report their average performance over three tasks for the final assessment. In our experiments, the dimension of all region embedding is set to 144.

\section{Results and Discussion}
Tables \ref{tab:NYC} and \ref{tab:CHI} show predictive performances on downstream tasks with and without income information. Models are sorted in descending order by the average coefficient of determination ($R^2$) among three tasks for each heterogeneous graph edge combination. \textit{Income + Mobility} achieves the best overall performance in both cities (ranked second by a narrow margin in Chicago). By adding income data, the $R^2$ values improved by 10.14 \%,  8.00\% and 10.22 \% in check-in, crime, and house price prediction in NYC and 9.00\% and 2.78\% in predicting crime and house price in CHI. Besides, results show that the performance of the heterogeneous graph learning model does not necessarily improve with more input data sources. 
 
 Echoing existing studies, we confirm mobility data as an important input component, as its absence results in a lower average $R^2$, largely attributed to the prediction of urban region popularity. However, large-scale mobility patterns are constructed on urban big data which is not always available and the processing of such data is also computationally expensive. Results show that for developing cities without access to fine-grained mobility data, \textit{Income + Neighbor} can effectively serve as an alternative solution or preliminary  estimation, with only a minor decrease in prediction accuracy, i.e., 11.82 \% and 9.42 \% in average $R^2$ for NYC and CHI, respectively. Moreover, this drop is primarily due to less accurate check-in predictions, while this combination can outperform \textit{Mobility + Neighbor +POI} in the other two tasks in CHI and house price prediction in NYC. More specifically, \textit{Income + Neighbor} measures the demographic and geographical nearby relationships between regions. Compared to integrating mobility data, this combination is slightly less capable of predicting mobility-based urban applications but can be more efficient for other downstream tasks.

 We evaluate the effect of including income data compared to the commonly used POI and geographic neighboring data by analyzing the improvement in model performance measured by the average testing $R^2$ resulting from the inclusion of each dimension in three prediction tasks. Income data increases the average testing $R^2$ by 0.143 and 0.103 respectively in NYC and CHI, compared with -0.05 and 0.005 for geographic proximity, and 0.037 and -0.083 for POI. While adding region income information consistently yields significantly better performance, adding both POI and geographic adjacency show varying contributions across different scenarios. 
 Despite that POI and geographic neighboring data are widely adopted in multi-view region representation learning, results show that adding POI and neighbor information can lead to poorer performance, especially when income data is already in use. 
 
 We extend the examination from income information to other easy-to-access demographic information, including age, education level, employment rate, and the percentage of foreigners. Table \ref{tab:otherfeatures} shows the $R^2$ by applying combinations of mobility and different demographic data to three downstream tasks. \textit{Income + Mobility} remains the best combination by multi-task average performance.
 Certain demographic information is effective to specific downstream tasks. For instance, \textit{Age + Mobility} achieved a 9.9\% higher $R^2$ value in check-in prediction in NYC compared with \textit{Mobility + Neighbor + POI}. \textit{Education + Mobility} achieved a 6.7\% higher $R^2$ in house price prediction in CHI. This implies that although demographic information can generally aid in learning region embedding, the extent of improvements by different attributes is context-aware and city-specific. This phenomenon also provides us with a glimpse into the differences and commonalities in segregation levels among various cities.

\begin{table*}[]
\caption{Performance comparison with different pre-train data combinations for three downstream tasks in New York City.}
\label{tab:NYC}
\begin{tabular}{lllllllllll}
\hline
\multirow{3}{*}{Combinations}      & \multicolumn{10}{c}{New York City (NYC)}                                                                                                                                         \\ \cline{2-11} 
                                   & \multicolumn{3}{c}{Check-in Prediction} & \multicolumn{3}{c}{Crime Rate Predition} & \multicolumn{3}{c}{House Price Prediction} & \multicolumn{1}{c}{\multirow{2}{*}{Average$R^2$}} \\ \cline{2-10}
                                   & MAE          & RMSE          & $R^2$& MAE         & RMSE        & $R^2$& MAE          & RMSE          & $R^2$& \multicolumn{1}{c}{
}                         \\ \hline
Income + Mobility                  
& 3484.555     & 9468.002      & 0.543    & 125.081     & 175.695     & 0.624    & 54.149       & 98.257        & 0.582       & 0.583                
\\ \hline
Income + Mobility + POI            
& 3694.521     & 9637.575      & 0.526    & 128.529     & 178.660     & 0.611    & 55.900       & 99.454        & 0.572       & 0.570                
\\ \hline
Income + Mobility + Neighbor + POI 
& 3939.863     & 9972.175      & 0.493    & 128.386     & 177.097     & 0.618    & 56.836       & 99.105        & 0.575       & 0.562                
\\ \hline
Income + Mobility + Neighbor       
& 4197.057     & 10202.192     & 0.470    & 130.899     & 180.523     & 0.603    & 58.294       & 101.958       & 0.550       & 0.541                
\\ \hline
Mobility + POI                     
& 3785.289     & 9978.571      & 0.493    & 136.262     & 186.103     & 0.578    & 62.871       & 103.427       & 0.537       & 0.536                
\\ \hline
Mobility + Neighbor + POI          
& 3949.923     & 9927.402      & 0.497    & 136.773     & 186.174     & 0.578    & 62.618       & 104.354       & 0.528       & 0.535                
\\ \hline
Mobility + Neighbor                
& 3992.503     & 10077.455     & 0.482    & 139.157     & 185.490     & 0.581    & 64.500       & 105.911       & 0.514       & 0.526                
\\ \hline
Mobility                           
& 3849.550     & 10464.986     & 0.442    & 140.252     & 193.084     & 0.546    & 58.619       & 100.691       & 0.561       & 0.516                
\\ \hline
Income + Neighbor                  
& 4915.606     & 11513.207     & 0.325    & 145.671     & 204.178     & 0.491    & 62.419       & 102.181       & 0.548       & 0.455                
\\ \hline
Income + Neighbor + POI            
& 4988.900     & 11752.269     & 0.296    & 138.520     & 195.215     & 0.536    & 61.631       & 104.161       & 0.530       & 0.454                
\\ \hline
Income + POI                       
& 4688.461     & 11502.841     & 0.326    & 157.457     & 213.123     & 0.447    & 72.020       & 107.033       & 0.504       & 0.426                
\\ \hline
Neighbor + POI                     & 5185.601     & 11896.113     & 0.279    & 159.190     & 213.502     & 0.444    & 74.307       & 115.005       & 0.427       & 0.383                \\ \hline
\end{tabular}
\end{table*}
\begin{table*}[]
\caption{Performance comparison with different pre-train data combinations  for three downstream tasks in Chicago.}
\label{tab:CHI}
\begin{tabular}{lllllllllll}
\hline
\multirow{3}{*}{Combinations}      & \multicolumn{10}{c}{Chicago (CHI)}                                                                                                                                         \\ \cline{2-11} 
                                   & \multicolumn{3}{c}{Check-in Prediction} & \multicolumn{3}{c}{Crime Rate Predition} & \multicolumn{3}{c}{House Price Prediction} & \multicolumn{1}{c}{\multirow{2}{*}{Average $R^2$}} \\ \cline{2-10}
                                   & MAE          & RMSE         & $R^2$& MAE         & RMSE        & $R^2$& MAE           & RMSE         & $R^2$& \multicolumn{1}{c}{}                         \\ \hline
Income + Mobility + Neighbor       & 3035.003     & 5687.108     & 0.659     & 237.607     & 328.937     & 0.789    & 19.732        & 25.745       & 0.890       & 0.779                                        \\ \hline
Income + Mobility                  & 2998.072     & 5726.868     & 0.655     & 228.956     & 335.273     & 0.781    & 21.264        & 26.967       & 0.881       & 0.772                                        \\ \hline
Income + Mobility + POI            & 3120.129     & 5826.669     & 0.643     & 236.191     & 344.476     & 0.768    & 20.321        & 26.809       & 0.882       & 0.764                                        \\ \hline
Mobility + Neighbor                & 2918.860     & 5456.713     & 0.686     & 263.013     & 376.547     & 0.724    & 22.052        & 28.486       & 0.866       & 0.759                                        \\ \hline
Income + Mobility + Neighbor + POI & 3052.954     & 5704.411     & 0.657     & 264.834     & 379.242     & 0.720    & 18.714        & 24.991       & 0.897       & 0.758                                        \\ \hline
Mobility                           & 2779.053     & 5384.743     & 0.695     & 256.442     & 359.376     & 0.747    & 25.698        & 33.225       & 0.819       & 0.754                                        \\ \hline
Mobility + POI                     & 2816.111     & 5557.032     & 0.675     & 279.561     & 402.229     & 0.685    & 24.881        & 32.267       & 0.829       & 0.730                                        \\ \hline
Mobility + Neighbor + POI          & 3057.951     & 5656.525     & 0.662     & 287.008     & 406.547     & 0.678    & 23.217        & 30.467       & 0.846       & 0.729                                        \\ \hline
Income + Neighbor                  & 4051.936     & 7033.644     & 0.478     & 291.682     & 388.259     & 0.706    & 22.867        & 28.787       & 0.864       & 0.683                                        \\ \hline
Income + Neighbor + POI            & 3869.772     & 6819.553     & 0.507     & 301.320     & 422.970     & 0.651    & 21.720        & 28.258       & 0.867       & 0.675                                        \\ \hline
Income + POI                       & 4202.467     & 7578.740     & 0.395     & 327.266     & 459.802     & 0.588    & 27.149        & 33.876       & 0.812       & 0.598                                        \\ \hline
Neighbor + POI                     & 3621.176     & 6386.388     & 0.570     & 435.984     & 590.534     & 0.321    & 34.041        & 42.737       & 0.697       & 0.529                                        \\ \hline
\end{tabular}
\end{table*}

\begin{table*}[]
\caption{A comparison of testing $R^2$ across various combinations of mobility and demographic data for pre-training. Best-performing combinations are highlighted in bold. }
\label{tab:otherfeatures}
\begin{tabular}{lllllllll}
\hline
\multirow{2}{*}{Combinations} & \multicolumn{2}{c}{Check-in Prediction} & \multicolumn{2}{c}{Crime Rate Predition} & \multicolumn{2}{c}{House Price Prediction} & \multicolumn{2}{c}{Average $R^2$}  \\ \cline{2-9} 
                              & NYC                & CHI                & NYC               & CHI              & NYC                  & CHI                 & NYC            & CHI            \\ \hline
Employment + Mobility         & 0.514              & 0.532              & 0.610             & 0.751            & 0.577                & 0.840               & 0.567          & 0.708          \\ \hline
Age + Mobility                & \textbf{0.546}     & 0.665              & 0.588             & 0.655            & 0.555                & 0.799               & 0.563          & 0.706          \\ \hline
Education + Mobility          & 0.495              & \textbf{0.672}     & 0.596             & 0.730            & 0.552                & \textbf{0.903}      & 0.548          & 0.768          \\ \hline
ForeignerBorn + Mobility      & 0.517              & 0.667              & 0.556             & 0.702            & 0.571                & 0.813               & 0.548          & 0.727          \\ \hline
Income + Mobility             & 0.543              & 0.655              & \textbf{0.624}    & \textbf{0.781}   & \textbf{0.582}       & 0.881               & \textbf{0.583} & \textbf{0.772} \\ \hline
\end{tabular}
\end{table*}

\section{Conclusion}
Demographic information is valuable for learning regional embedding. Income level describes the region's inherent attributes while mobility data captures the inter-region interaction. Our experiments on predicting regional check-in counts, crime rate, and house prices in both New York and Chicago confirm the importance of fine-grained mobility data in learning region embedding and show that with additional demographic information, \textit{Income + Mobility} improves the state-of-the-art prediction accuracy by up to 10.22\%. For developing countries without access to mobility data, we suggest \textit{geographic proximity + income} as an alternative combination of pre-train data for generating regional embedding. This work demonstrates the potential of region embedding to enable transferable urban prediction models.

\bibliographystyle{ACM-Reference-Format}
\bibliography{sample-base}

\appendix

\end{document}